\documentclass[a4paper,color,twoside,amssymb,twocolumn]{article}
\usepackage{cite}
\usepackage{Latex-document}
\usepackage{bm}
\usepackage[usenames]{color}
\definecolor{lightblue}{rgb}{0, 0, 0}

\usepackage{geometry}
\usepackage{epstopdf}
\usepackage{listings}
\usepackage{url}
\usepackage{color}
\usepackage{xcolor}
\usepackage{framed}
\usepackage{enumerate}
\usepackage{amsmath}
\usepackage{graphicx}  
\usepackage{times}  
\usepackage{helvet}  
\usepackage{courier}  
\usepackage{footnote}
\usepackage{lipsum}
\usepackage[justification=centering]{caption}

\definecolor{codegreen}{rgb}{0,0.6,0}
\definecolor{codegray}{rgb}{0.5,0.5,0.5}
\definecolor{codepurple}{rgb}{0.58,0,0.82}
\definecolor{backcolour}{rgb}{0.95,0.95,0.92}
\lstdefinestyle{mystyle}{
	commentstyle=\color{codegreen},
	keywordstyle=\color{magenta},
	numberstyle=\color{codegray},
	stringstyle=\color{codepurple}\ttfamily,
	basicstyle=\ttfamily\small,
	breakatwhitespace=false,         
	breaklines=true,                 
	captionpos=b,                    
	keepspaces=true,                 
	numbers=left,                    
	numbersep=-5pt,                  
	showspaces=false,                
	showstringspaces=false,
	showtabs=false,                  
	tabsize=4
}

\newcommand{\tabincell}[2]{\begin{tabular}{@{}#1@{}}#2\end{tabular}}

\newcommand{\titlename}{Code Attention: Translating Code to Comments by Exploiting Domain Features}
\newcommand{\authorname}{Wenhao Zheng, Hong-Yu Zhou, Ming Li, Jianxin Wu}
\newcommand{\atime}{Received month dd.yyyy; accepted month dd.yyyy}
\newcommand{\aemail}{$lim@nju.edu.cn$}

\begin{document}

\thispagestyle{first}
\setcounter{page}{1}

\begin{tabular*}{\textwidth}{l}
 \hspace*{-6.1mm}\includegraphics{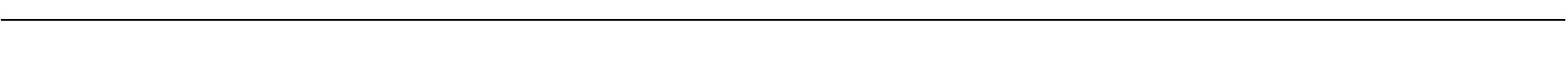}\vspace{-6.4mm}\\
 \hspace*{-6mm}\colorbox{lightblue}{
\arraycolsep=132pt \normalsize\hspace*{-10mm}{\color[cmyk]{.0, 0.0, 0, .0}$\begin{array}{l}\\[-3.5mm]\bf \hspace*{-37mm}
RESEARCH~ARTICLE
\end{array}$}}\vspace{-2mm}\\
\end{tabular*}

\newcommand\blfootnote[1]{%
	\begingroup 
	\renewcommand\thefootnote{}\footnote{#1}%
	\addtocounter{footnote}{-1}%
	\endgroup 
}

\begin{strip}
\begin{center}
{\tfont \LARGE \titlename}\\[6mm]
{\bf \authorname}\\[3mm]
\normalsize{National Key Laboratory for Novel Software Technology, Nanjing University, Nanjing 210023, China}
\end{center}
\cnote
\end{strip}

\Abstract{Appropriate comments of code snippets provide insight for code functionality, which are helpful for program comprehension. However, due to the great cost of authoring with the comments, many code projects do not contain adequate comments. Automatic comment generation techniques have been proposed to generate comments from pieces of code in order to alleviate the human efforts in annotating the code. Most existing approaches attempt to exploit certain correlations (usually manually given) between code and generated comments, which could be easily violated if the coding patterns change and hence the performance of comment generation declines. Furthermore, previous datasets are too small to validate the methods and show their advantage. In this paper, we first build C2CGit, a large dataset from open projects in GitHub, which is more than 20$\times$ larger than existing datasets. Then we propose a new attention module called Code Attention to translate code to comments, which is able to utilize the domain features of code snippets, such as symbols and identifiers. By focusing on these specific features, Code Attention has the ability to understand the structure of code snippets. Experimental results demonstrate that the proposed module has better performance over existing approaches in both BLEU, METEOR and human evaluation. We also perform ablation studies to determine effects of different parts in Code Attention.}

\blfootnote{\footname}\\
\renewcommand{\thefootnote}{\fnsymbol{footnote}}\footnote{}


\section{Introduction}
\label{sec:introduction}

\noindent Program comments usually provide insight for code functionality, which are important for program comprehension, maintenance and reusability. For example, comments are helpful  for working efficiently in a group or integrating and modifying open-source software. However, because it is time-consuming to create and update comments constantly, plenty of source code, especially the code from open-source software, lack adequate comments~\cite{fluri2007code}. Source code without comments would reduce the maintainability and usability of software.

To mitigate the impact, automatic program annotation techniques have been proposed to automatically supplement the missing comments by analyzing source code.~\cite{sridhara2010towards} generated summary comments by using variable names in code.~\cite{rastkar} managed to give a summary by reading software bug reports.~\cite{mcburney2014automatic} leveraged the documentation of API to generate comments of code snippets.

As is well known, source code are usually structured while the comments in natural language are organized in a relatively free form. Therefore, the key in automatic program annotation is to identify the relationship between the functional semantics of the code and its corresponding textual descriptions. Since identifying such relationships from the raw data is rather challenging due to the heterogeneity nature of programming language and natural language, most of the aforementioned techniques usually rely on certain assumptions on the correlation between the code and their corresponding comments (e.g., providing paired code and comment templates to be filled in), based on which the code are converted to comments in natural language. However, the assumptions may highly be coupled with certain projects while invalid on other projects. Consequently, these approaches may have large variance in performances on real-world applications.

In order to improve the applicability of automatic code commenting, machine learning has been introduced to learn how to generate comments in natural language from source code in programming languages.~\cite{srinivasan2016summarizing} and~\cite{cnnforsummarization} treated source code as natural language texts, and learned a neural network to summarize the words in source code into briefer phrases or sentences. However, as pointed out by~\cite{huo2016learning}, source code carry non-negligible semantics on the program functionality and should not be simply treated as natural language texts. Therefore, the comments generated by~\cite{srinivasan2016summarizing} may not well capture the functionality semantics embedded in the program structure. For example, as shown in Figure \ref{code0}, if only considering the lexical information in this code snippet, the comment would be ``\emph{swap two elements in the array}''. However, if considering both the structure and the lexical information, the correct comment should be ``\emph{shift the first element in the array to the end}''.
\begin{figure}[htpb]
\begin{lstlisting}[language=Java]
  int i = 0;
  while(i<n){
	// swap is a build-in function in Java
	swap(array[i],array[i+1]);
    i++;}\end{lstlisting}
	\caption{An example of code snippet. If the structural semantics provided by the \texttt{while} is not considered, comments indicating wrong semantics may be generated.}
	\label{code0}
\end{figure}

\begin{figure}[!t]
	\centering
	{\includegraphics[width=1\columnwidth]{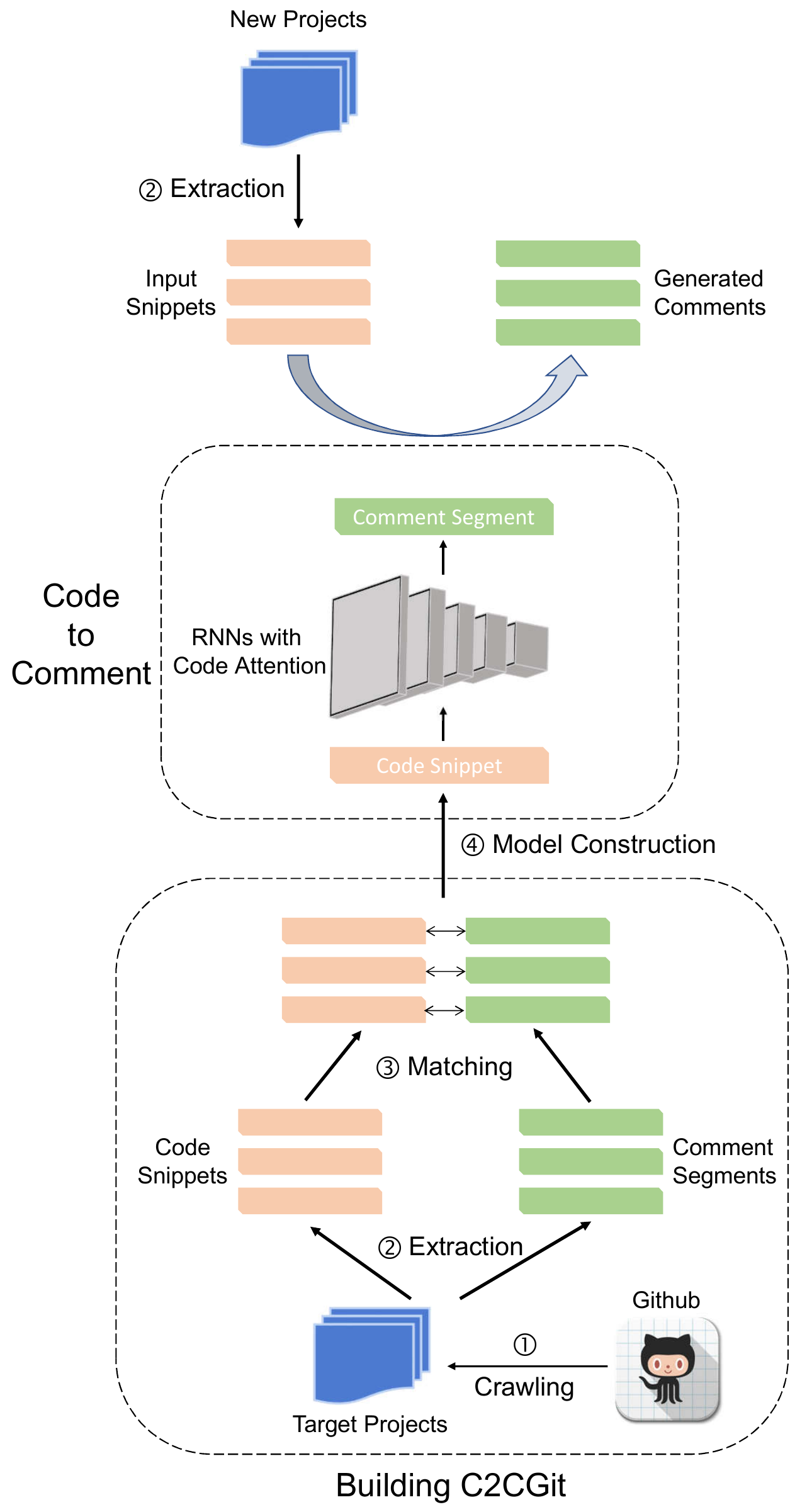} }
	\caption{The whole framework of our proposed method. The main skeleton includes two parts: building C2CGit and code to comments translation.}
	\label{framework}
\end{figure}

One question arises: Can we \emph{directly learn a mapping} between two heterogeneous languages? Inspired by the recent advances in neural machine translation (NMT), we propose a novel attention mechanism called Code Attention to directly \emph{translate} the source code in programming language into comments in natural language. Our approach is able to explore domain features in code by attention mechanism, e.g. explicitly modeling the semantics embedded in program structures such as loops and symbols, based on which the functional operations of source code are mapped into words. To verify the effectiveness of Code Attention, we build C2CGit, a large dataset collected from open source projects in Github. The whole framework of our proposed method is as shown in Figure \ref{framework}. Empirical studies indicate that our proposed method can generate better comments than previous work, and the comments we generate would conform to the functional semantics in the program, by explicitly modeling the structure of the code.

The rest of this paper is organized as follows. After briefly introducing the related work and preliminaries, we describe the process of collecting and preprocessing data in Section 4, in which we build a new benchmark dataset called C2CGit.  In Section 5, we introduce the Code Attention module, which is able to leverage the structure of the source code. In Section 6, we report the experimental results by comparing it with five popular approaches against different evaluation metrics. On BLEU and METEOR, our approach outperforms all other approaches and achieves new state-of-the-art performance in C2CGit.

Our contribution can be summarized as:
\begin{enumerate}[i)]
	\item A new benchmark dataset for code to comments translation. C2CGit contains over 1k projects from GitHub, which makes it more real and 20$\times$ larger than previous dataset~\cite{srinivasan2016summarizing}.
	
	\item We explore the possibility of whether recent pure attention model~\cite{attention} can be applied to this translation task. Experimental results show that the attention model is inferior to traditional RNN, which is the opposite to the performance in NLP tasks.
	
	\item To utilize domain features of code snippets, we propose a Code Attention module which contains three steps to exploit the structure in code. Combined with RNN, our approach achieves the best results on BLEU and METEOR over all other methods in different experiments.
\end{enumerate}

\section{Related Work}

Previously, there already existed some work on producing code descriptions based on source code. These work mainly focused on how to extract key information from source code, through rule-based matching, information retrieval, or probabilistic methods.~\cite{sridhara2010towards} generated conclusive comments of specific source code by using variable names in code.~\cite{sridhara2011automatically} used several templates to fit the source code. If one piece of source code matches the template, the corresponding comment would be generated automatically.~\cite{movshovitz} predicted class-level comments by utilizing open source Java projects to learn n-gram and topic models, and they tested their models using a character-saving metric on existing comments.
There are also retrieval methods to generate summaries for source code based on automatic text summarization~\cite{haiduc2010use} or topic modeling~\cite{eddy2013evaluating}, possibly combining with the physical actions of expert engineers~\cite{rodeghero2014improving}.

\textbf{Datasets}. There are different datasets describing the relation between code and comments. Most of datasets are from Stack Overflow~\cite{boa,wong2013autocomment,srinivasan2016summarizing} and GitHub~\cite{wong2015clocom}. Stack Overflow based datasets usually contain lots of pairs in the form of Q\&A, which assume that real world code and comments are also in Q\&A pattern. However, this assumption may not hold all the time because those questions are carefully designed. On the contrary, we argue that current datasets from GitHub are more real but small, for example,~\cite{wong2015clocom} only contains 359 comments. In this paper, our C2CGit is much larger and also has the ability to keep the accuracy. 

\textbf{Machine Translation}. In most cases, generating comments from source code is similar to the sub-task named machine translation in natural language processing (NLP). There have been many research work about machine translation in this community.~\cite{mathematicsofml} described a series of five statistical models of the translation process and developed an algorithm for estimating the parameters of these models given a set of pairs of sentences that each pair contains mutual translations, and they also define a concept of word-by-word alignment between such pairs of sentences. 
\cite{koehn2003statistical} proposed a new phrase-based translation model and decoding algorithm that enabled us to evaluate and compare several previously proposed phrase-based translation models. However, the system itself consists of many small sub-components and they are designed to be tuned separately. Although these approaches achieved good performance on NLP tasks, few of them have been applied on code to comments translation. Recently, deep neural networks achieve excellent performance on difficult problems such as speech recognition~\cite{dlforspeech1}, visual object recognition~\cite{alexnet} and machine translation~\cite{seq2seq}. For example, the neural translator proposed in~\cite{seq2seq} is a newly emerging approach which attempted to build and train a single, large neural network which takes a sentence as an input and outputs a corresponding translation.

Two most relevant works are~\cite{srinivasan2016summarizing} and~\cite{cnnforsummarization}.~\cite{cnnforsummarization} mainly focused on extreme summarization of source code snippets into short, descriptive function name-like
summaries but our goal is to generate human-readable comments of code snippets.~\cite{srinivasan2016summarizing} presented the first completely data driven approach for generating high level summaries of source code by using Long Short Term Memory (LSTM) networks to produce sentences. However, they considered the code snippets as natural language texts and employed roughly the same method in NLP without considering the structure of code. 

Although translating source code to comments is similar to language translation, there does exist some differences. For instance, the structure of code snippets is much more complex than that of natural language and usually has some specific features, such as various identifiers and symbols; the length of source code is usually much longer than the comment; some comments are very simple while the code snippets are very complex. All approaches we have mentioned above do not make any optimization for source code translation. In contrast, we design a new attentional unit called Code Attention which is specially optimized for code structure to help make the translation process more specific. By separating the identifiers and symbols from natural code segments, Code Attention is able to understand the code snippets in a more structural way. 

\section{Preliminaries}
\label{preliminaries}

In this section, we introduce the recurrent neural networks (RNNs), a family of neural networks designed for processing sequential data. Some traditional types of neural networks (e.g., convolution neural networks, recursive networks) make an assumption that all elements are independent of each other, while RNNs perform the same task with the output being depended on the previous computations. For instance, in natural language processing, if you want to predict the next word in a sentence you better know which words come before it. 

\subsection{seq2seq model}
A recurrent neural network (RNN) is a neural network that consists of a hidden state \textbf{h} and an optional output \textbf{y} which operates on a variable length sequence. An RNN is able to predict the next symbol in a sequence by modeling a probability distribution over the sequence \textbf{x}$=(x_1, \dots, x_T)$. At each timestep $t$, the hidden state $\textbf{h}_t$ is updated by
\begin{equation}
\label{fm1}
\textbf{h}_t=f_{encoder}(\textbf{h}_{t-1}, \textbf{x}_t)
\end{equation}
where $f_{encoder}$ is a non-linear activation function (e.g., sigmoid function \cite{yin2003flexible}, LSTM~\cite{lstms}, GRU~\cite{grus}). One usual way of defining the recurrent unit $f_{encoder}$ is a linear transformation plus a nonlinear activation, e.g., 
\begin{equation}
\label{activation}
\textbf{h}_t=tanh(W[\textbf{h}_{t-1}, \textbf{x}_t]+\textbf{b})
\end{equation}
where we parameterized the relation between $\textbf{h}_{t-1}$ and $\textbf{x}_t$ into matrix $W$, and $\textbf{b}$ is the bias term. Each element of its input is activated by the function $tanh$. A simple RNN aims to learn the parameters $W$ and $\textbf{b}$. In this case, we can get the final joint distribution, 
\begin{equation}
\label{distribution}
p(\textbf{x})=\prod_{T}^{t=1}p(\textbf{x}_t|\textbf{x}_1,\dots,\textbf{x}_{t-1})
\end{equation}

The basic cell unit in RNN is important to decide the final performance. A gated recurrent unit is proposed by Cho et al.~\cite{GRU} to make each recurrent unit to
adaptively capture dependencies of different time scales. GRU has gating units but no separate memory cells when compared with LSTM.

\begin{figure}[!t]
	\centering
	{\includegraphics[width=0.8\columnwidth]{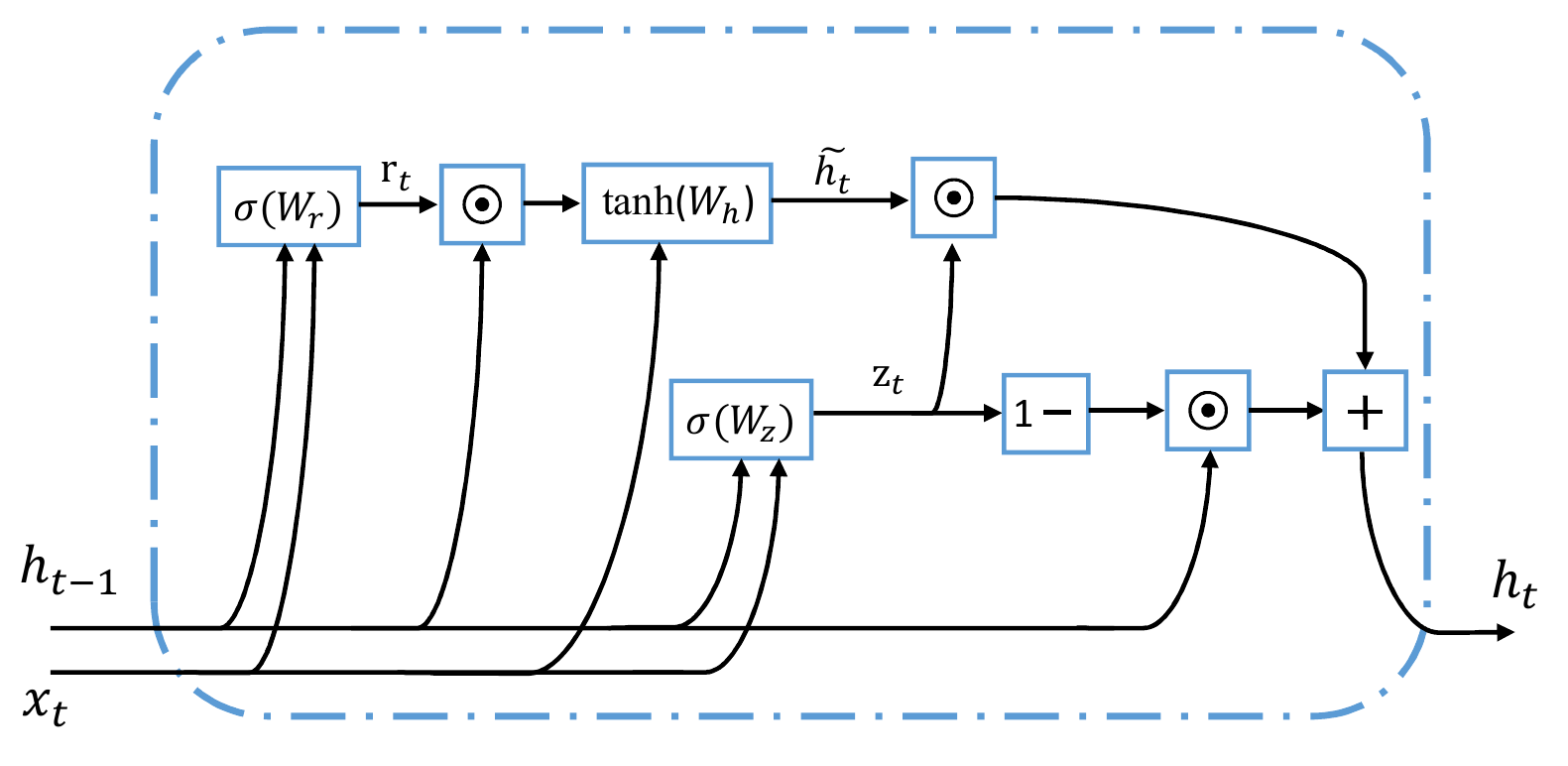} }
	\caption{The data flow and operations in GRU.}
	\label{GRU}
\end{figure}

GRU contains two gates: an update gate $\textbf{z}$ and a reset gate $\textbf{r}$ which correspond to forget gate and input gate, respectively. We show the update rules of GRU in the Equations (4) to (7),
\begin{eqnarray}
\label{GRU_formula}
\textbf{z}_t&=& \sigma(W_z[\textbf{h}_{t-1}, \textbf{x}_t]+\textbf{b}_z)\\
\textbf{r}_t&=& \sigma(W_r[\textbf{h}_{t-1}, \textbf{x}_t]+\textbf{b}_r)\\
\widetilde{\textbf{h}}_t&=& tanh(W_h[\textbf{r}_{t}\odot\textbf{t}_{t-1}, \textbf{x}_t]+\textbf{b}_h)\\
\textbf{h}_t&=& (1-\textbf{z}_t)\odot\textbf{h}_{t-1}+\textbf{z}_{t}\odot\widetilde{\textbf{h}}_t
\end{eqnarray}
where $\sigma(x) = \frac{1}{1+\exp(-x)}$, $\circ$ is the component-wise product
between two vectors. For a better understanding, we also provide the data flow and operations in Figure~\ref{GRU}. There are two reasons which make us choose GRU: the first one is that Chung et al.~\cite{Evaluation} found that when LSTM and GRU have the same amount of parameters, GRU slightly outperforms LSTM; the second is that GRU is much easier to implement and train compared with LSTM.

In order to learn a better phrase representations, a classical recurrent neural network architecture learns to encode a variable-length inputs into a fixed-length vector representation and then to decode the vector into a variable-length output. To be simple, this architecture bridges the gap between two variable-length vectors. While if we look inside the architecture from a more probabilistic perspective, we can rewrite Eq. (\ref{distribution}) into a more general form, e.g., $p(y_1, \dots, y_K\ |\ x_1, \dots, x_T)$, where it is worth noting that the length of input and output may differ in this case.

Above model contains two RNNs. The first one is the encoder, while the other is used as a decoder. The encoder is an RNN that reads each symbol of an input sequence \textbf{x} sequentially. As it reads each symbol, the hidden state of the encoder updates according to Eq. (\ref{fm1}). At the end of the input sequence, there is always a symbol telling the end, and after reading this symbol, the last hidden state is a summary \textbf{c} of the whole input sequence.

As we have discussed, the decoder is another RNN which is trained to generate the output sequence by predicting the next symbol $\textbf{y}_t$ given the hidden state $\textbf{h}_t$.
\begin{equation}
p(\textbf{y}_t\ |\ \textbf{y}_{t-1},\dots,\textbf{y}_1, \textbf{c}) = f_{decoder}(\textbf{h}_t, \textbf{y}_{t-1}, \textbf{c}),
\end{equation}
where $\textbf{h}_t = f(\textbf{h}_{t-1}, \textbf{y}_{t-1}, \textbf{c})$ and $f_{decoder}$ is usually a softmax function to produce valid probabilities. Note that there are several differences between this one and the original RNN. The first is the hidden state at timestep $t$ is no longer  based on $\textbf{x}_{t-1}$ but on the $\textbf{y}_{t-1}$ and the summary $\textbf{c}$, and the second is that we model $\textbf{y}_t$ and $\textbf{x}_t$ jointly which may result in a better representation.

\subsection{Attention Mechanism}
A potential issue with the above encoder-decoder approach is that a recurrent neural network has to compress all the necessary information of $x_1,\ \dots,\ x_T$ into a context vector $\textbf{c}$ for all time, which means the length of vector $\textbf{c}$ is fixed. There are several disadvantages here. This solution may make it difficult for the neural network to cope with long sentences, especially those that are longer than the sentences in the training corpus, and Cho~\cite{SMT} showed that indeed the performance of a basic encoder–decoder deteriorates rapidly as the length of an input sentence increases. Specifically, when backing to code-to-comment case, every word in the code may have different effects on each word in the comment. For instance, some \emph{keywords} in the source code can have direct influences on the comment while others do nothing to affect the result.

\begin{figure*}[!t]
	\centering
	{\includegraphics[width=1.8\columnwidth]{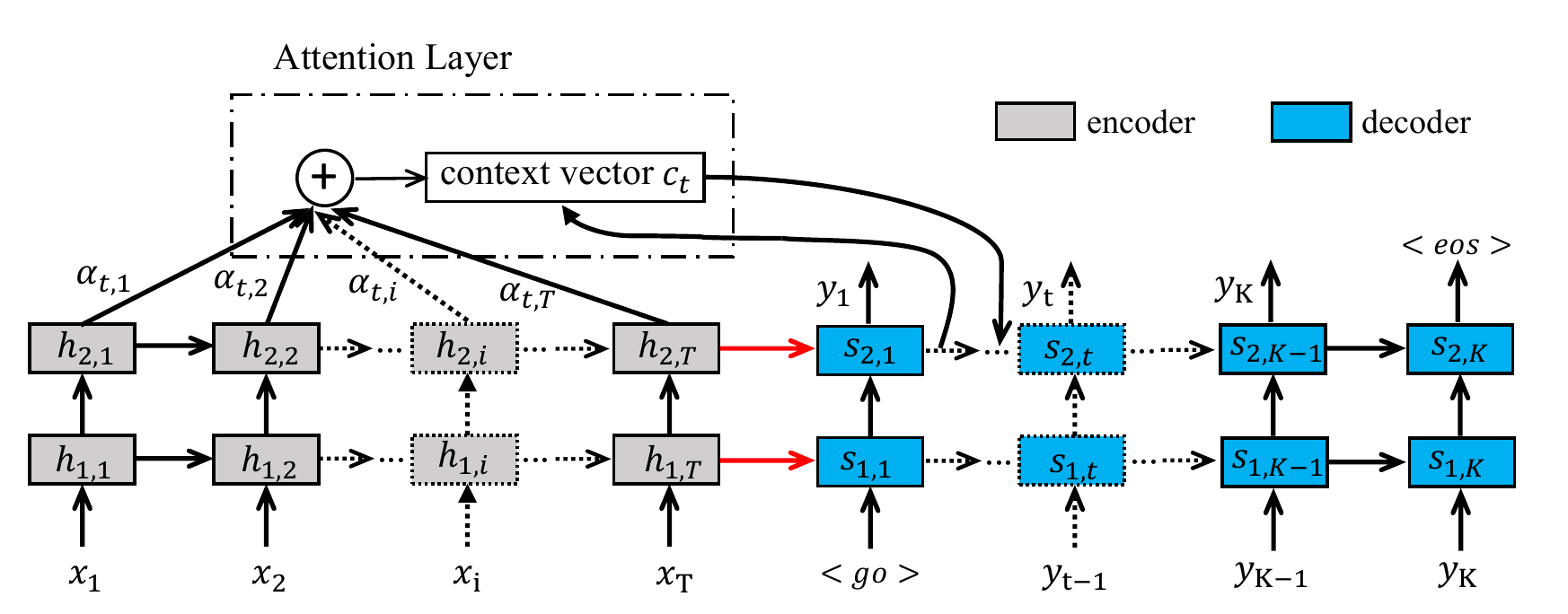} }
	\caption{An overview of the translation model. We employ a two-layer recurrent model, where the gray box represents the encoder unit and the blue ones represent the decoder part.}
	\label{seq2seq}
\end{figure*}

Considering all factors we have talked above, a global attention mechanism should be existed in a translation system. An overview of the model is provided in Fig.~\ref{seq2seq}. $\textbf{h}_{i,j}$ is the hidden state located at the $i$th ($i=1,2$) layer and $j$th ($j=1,\dots,T$) position in the encoder. $\textbf{s}_{i,k}$ is the hidden state located at the $i$th ($i=1,2$) layer and $j$th ($k=1,\dots,K$) position in the decoder. Instead of LSTM, GRU~\cite{GRU} could be used as the cell of both $f_{encoder}$ and $f_{decoder}$. Unlike the fixed vector $\textbf{c}$ in the traditional encoder-decoder approach, current context vector $\textbf{c}_t$ varies with the step $t$, 
\begin{equation}
\textbf{c}_t=\sum_{j=1}^{T}\alpha_{t,j}{\textbf{h}_{2,j}}
\end{equation}
and then we can get a new form of $\textbf{y}_t$,
\begin{equation}
\textbf{y}_t=f_{decoder}(\textbf{c}_t, \textbf{s}_{2,t-1}, \textbf{s}_{1,t})
\end{equation}
where $\alpha_{t,j}$ is the weight term of $j$th location at step $t$ in the input sequence. Note that the weight term $\alpha_{t,j}$ is normalized to $\left[0,1\right]$ using a softmax function,
\begin{equation}
\alpha_{t,j}=\frac{\exp(\textbf{e}_{t,j})}{\sum_{i=1}^{T}\exp(\textbf{e}_{t,i})},
\end{equation}
where $\textbf{e}_{t,j}=a(\textbf{s}_{2,t-1}, \textbf{h}_{2,j})$ scores how well the inputs around position $j$ and the output at position $t$ match and is a learnable parameter of the model.



\section{C2CGit: A New Benchmark for Code to Comment Translation}\label{dataset}

For evaluating proposed methods effectively, we build the C2CGit dataset firstly. We collected data from GitHub, a web-based Git repository hosting service. We crawled over 1,600 open source projects from GitHub, and got 1,006,584 Java code snippets. After data cleaning, we finally got \textbf{879,994} Java code snippets and \textbf{the same number of} comment segments. Although these comments are written by different developers with different styles, there exist common characteristics under these styles. For example, the exactly same code could have totally different comments but they all explain the same meaning of the code. In natural language, same source sentence may have more than one reference translations, which is similar to our setups. We name our dataset as \textbf{C2CGit}.

\begin{figure}[!tpb]
	\centering
	\footnotesize
	{\includegraphics[width=1\columnwidth]{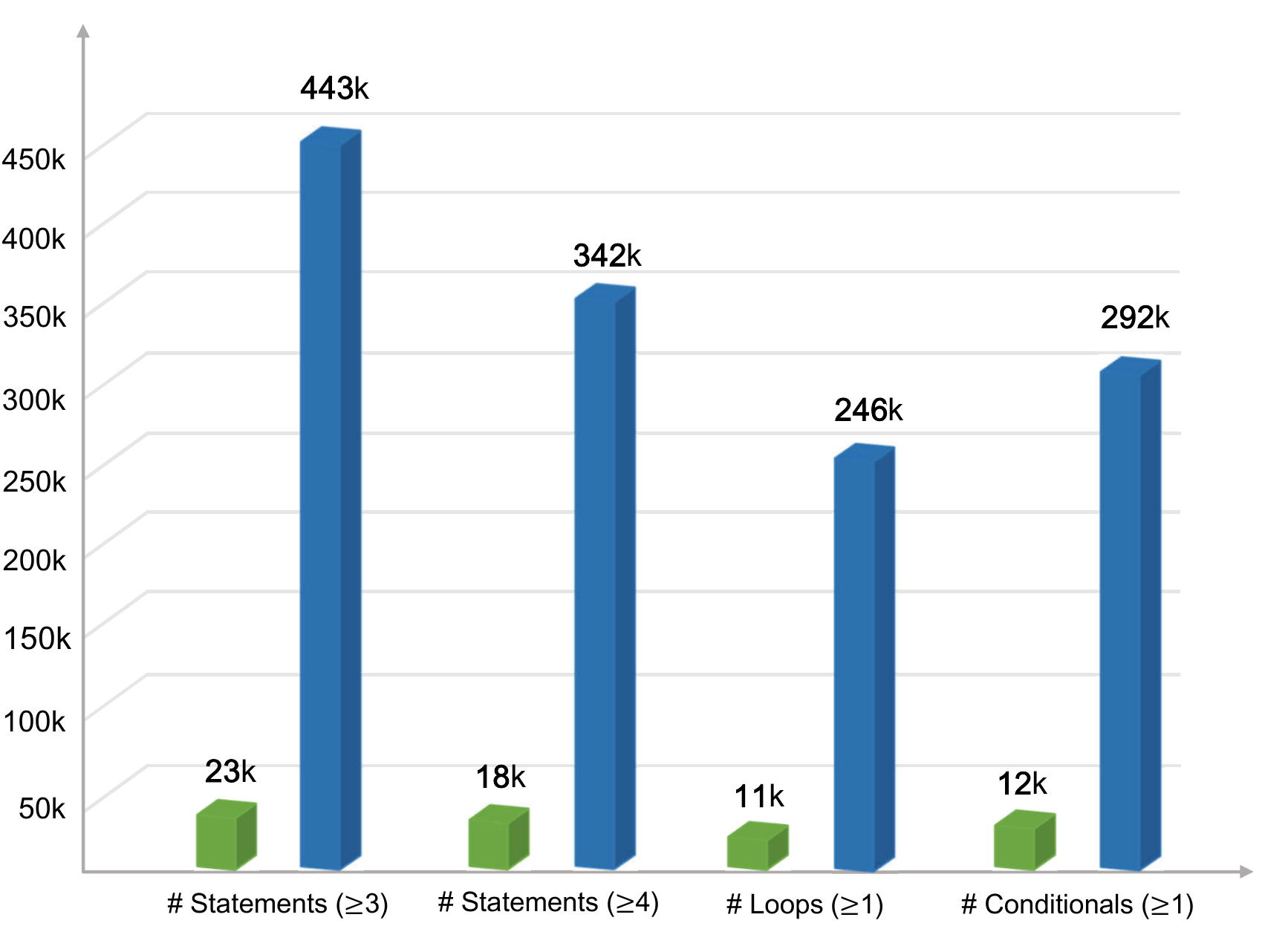} }
	\caption{We make a comparison between C2CGit (blue) and CODE-NN (green). We can see that C2CGit is larger and more diversified.}
	\label{hist}
\end{figure}

To the best of our knowledge, there does not exist such a large public dataset for code and comment pairs. One choice is using human annotation~\cite{oda2015learning}. By this way, the comments could have high accuracy and reliability. However, it needs many experienced programmers and consumes a lot of time if we want to get big data.  Another choice is to use recent CODE-NN~\cite{srinivasan2016summarizing} which mainly collected data from Stack Overflow which contains some code snippets in answers. For the code snippet from accepted answer of one question, the title of this question is regarded as a comment. Compared with CODE-NN (C$\#$), our C2CGit (Java) holds two obvious advantages:
\begin{itemize}
	\item \textbf{Code snippets in C2CGit are more real}. In many real projects from C2CGit, several lines of comments often correspond to a much larger code snippet, for example, a 2-line comment is annotated above 50-line code. However, this seldom appears in Stack Overflow.
	\item \textbf{C2CGit is much larger and more diversified than CODE-NN}. We make a detailed comparison in Figure~\ref{hist} and Table~\ref{length}. We can see that C2CGit is about 20$\times$ larger than CODE-NN no matter in statements, loops or conditionals. Also, C2CGit holds more tokens and words which demonstrate its diversity.
\end{itemize}

\textbf{Extraction}. We downloaded projects from the GitHub website by using web crawler.\footnote{The crawler uses the Scrapy framework. Its documentation can be found in \url{http://scrapy.org}} Then, the Java file can be easily extracted from these projects. Source code and comments should be split into segments. If we use the whole code from a Java file as the input and the whole comments as the output, we would get many long sentences and it is hard to handle them both in statistical machine translation and neural machine translation. Through analyzing the abstract syntax tree (AST)~\cite{neamtiu2005understanding} of code, we got code snippets from the complete Java file. By leveraging the method raised by~\cite{wong2015clocom}, the comment extraction is much easier, since it only needs to detect different comment styles in Java.

\begin{table}[!t]
	
	\footnotesize
	\setlength{\tabcolsep}{3pt} 
	\caption{Average code and comments together with vocabulary sizes for C2CGit, compared with CODE-NN.}
	\centering
	\label{length}
	\begin{tabular}{c|c|c|c|c}
		&Avg. code length & Avg. title length & tokens & words \\
		\hline
		\hline
		\textbf{CODE-NN} & 38 tokens & 12 words & 91k & 25k\\
		\hline
		\textbf{C2CGit} & \textbf{128 tokens} & \textbf{22 words} & \textbf{129,340k} & \textbf{22,299k}\\ 
		\hline
	\end{tabular}
\end{table}

\textbf{Matching}. Through the above extraction process, one project would generate many code snippets and comment segments. The next step is to find a match between code snippets and comment segments. We extracted all identifiers other than keyword nodes from the AST of code snippets. Besides, the Java code prefer the camel case convention (e.g., StringBuilder can be divided into two terms, string and builder). Each term from code snippets is then broken down based on the camel case convention. Otherwise, if a term uses underline to connect two words, it can also be broken down. After these operations, a code snippet is broken down to many terms. Because comments are natural language, we use a tokenization tool\footnote{http://www.nltk.org/}, widely used in natural language processing to handle the comment segments. If one code snippet shares the most terms with another comment segment, the comment segment can be regarded as a translation matching to this code snippet.

\textbf{Cleaning}. We use some prior knowledge to remove noise in the dataset. The noise is from two aspects. One is that we have various natural languages, the other is that the shared words between code snippets and comment segments are too few. Programmers coming from all around the world can upload projects to GitHub, and their comments usually contain non-English words. These comments would make the task more difficult but only occupy a small portion. Therefore, we deleted instances containing non-English words (non-ASCII characters) if they appear in either code snippets or comment segments. Some code snippets only share one word or two with comment segments, which suggests the comment segment can't express the meaning of code. These code and comment pairs also should be deleted.

\section{Proposed method: Code Attention Mechanism}
\label{cd}

\begin{figure*}[ht]
	\centering
	{\includegraphics[width=2\columnwidth]{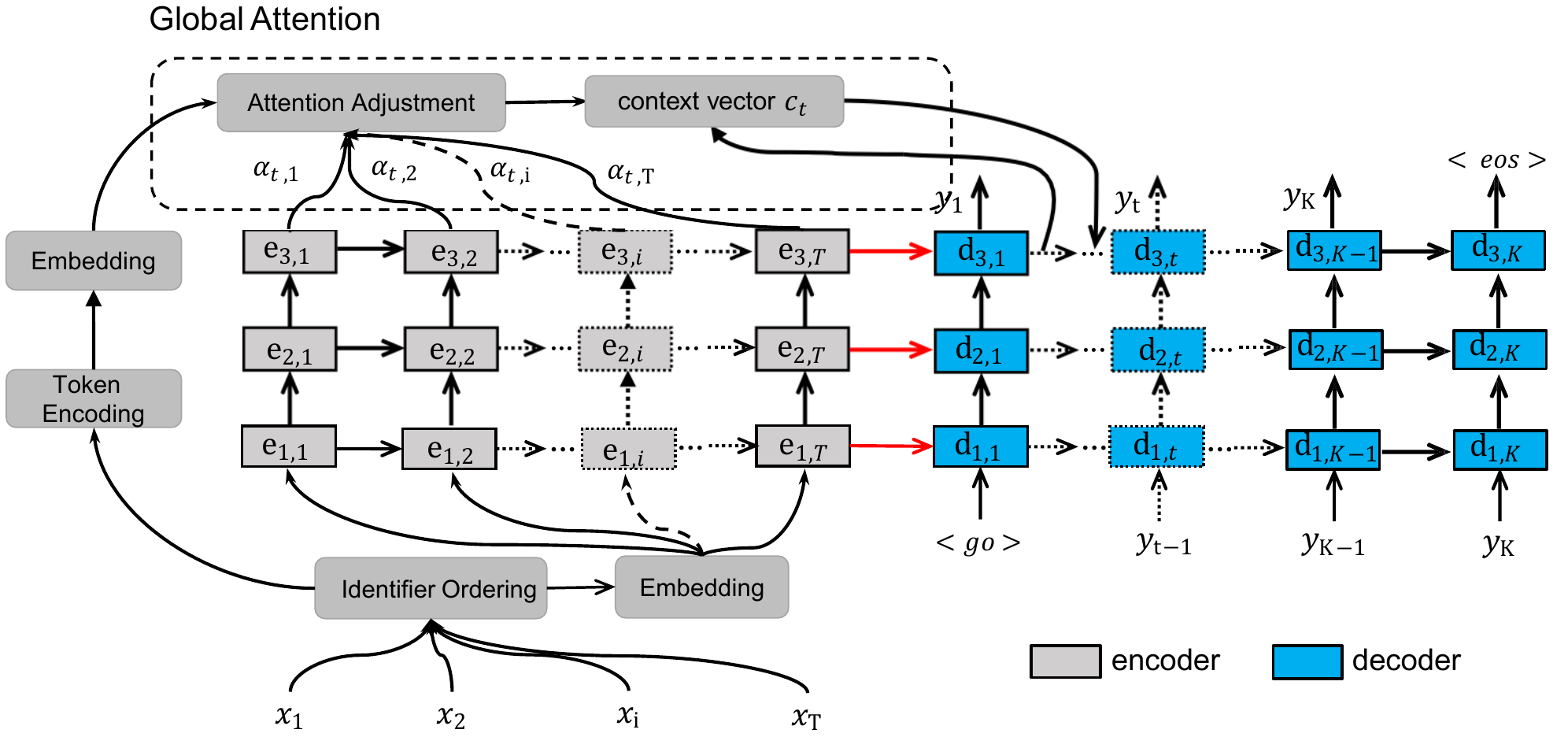} }
	\caption{The whole model architecture. Note that Code Attention mainly contains 3 steps: Identifier Ordering, Token Encoding and Global Attention. The first two module are followed by two independent embedding layers as shown in the flow diagram above.}
	\label{attention}
\end{figure*}

In this section, we mainly talk about the Code Attention mechanism in the model. For the encoder-decoder structure, we first build a 3-layer translation model as Section 3 said, whose basic element is Gated Recurrent Unit (GRU). Then, we modify the classical attention module in encoder. To be specific, we consider the embedding of symbols in code snippets as learnable prior weights to evaluate the importance of different parts of input sequences.
For convenience, we provide an overview of the entire model in Figure~\ref{attention}.

Unlike traditional statistical language translation, code snippets have some different characteristics, such as some identifiers ({\em{for}} and {\em{if}}) and different symbols (e.g., $\times, \div,=$). However, former works usually ignore these differences and employ the common encoding methods in NLP. In order to underline these identifiers and symbols, we simply import two strategies: Identifier Ordering and Token Encoding, after which we then develop a Global Attention module to learn their weights in input code snippets. We will first introduce details of Identifier Ordering and Token Encoding in the following.

\textbf{Identifier Ordering}. As the name suggests, we directly sort {\em{for}} and {\em{if}} in code snippets based on the order they appear. After sorting, 
\begin{eqnarray}
for/if \longrightarrow for/if + N \nonumber
\end{eqnarray}
where N is decided by the order of each identifier in its upper nest. For example, when we have multiple {\em if} and {\em for}, after identifier sorting, we have such forms,

\begin{figure}[!htp]
	\begin{lstlisting}[language=JAVA,morekeywords={FOR1,FOR2,IF1,IF2,ENDIF1,ENDIF2,ENDFOR1,ENDFOR2}]
  FOR1(i=0; i<len - 1; i++)
    FOR2(j=0; j<len - 1 - i; j++)
      IF1(arr[j] > arr[j + 1])
	    temp = arr[j]
	    arr[j] = arr[j+1]
	    arr[j+1] = temp
	  ENDIF1
	ENDFOR2
  ENDFOR1\end{lstlisting}
	\caption{An example of collected code snippet after identifier sorting.}
	\label{idsort}
\end{figure}
We can see that replaced identifiers are able to convey the original order of each of them. It is worth noting that Identifier Ordering makes a difference among \em{fors} or \em{ifs} appeared in different loop levels. 

\textbf{Token Encoding}. In order to stress the distinction among tokens e.g. symbols, variables and keywords in code snippets, these tokens should be encoded in a way which helps make them more conspicuous than naive encoded inputs. To be specific, we first build a dictionary including all symbols, like $\times,\ \div,\ ;,\ \left\{,\ \right\}$ and keywords, such as $int, float, public, ...$  in code snippets. The tokens not containing in this dictionary are regarded as variables. Next, we construct an independent token vocabulary which is the same size as the vocabulary of all input snippets, and encode these tokens using an extra embedding matrix. The embedded tokens can be treated as learnable weights in Global Attention.

\subsection{Global Attention}
In order to underline the importance of symbols in code, we import a novel attention mechanism called Global Attention. We represent $\textbf{x}$ as a set of inputs. Let $Ident(\cdot)$ and $Sym(\cdot)$ stand for our \emph{Identifier ordering} and \emph{Token Encoding}, respectively. $E(\cdot)$ be used to represent the embedding method. The whole Global Attention operation can be summarized as,
\begin{eqnarray}
E(Sym\left(Ident\left(\textbf{x}\right)\right))\ \textcircled{$\times$}\ f_{e}(\textbf{x})
\end{eqnarray}
where $f_e(\cdot)$ is the encoder, \textcircled{$\times$} represents dot product to stress the effects of encoded tokens.

After Token Encoding, we now have another token embedding matrix: $\textbf{F}$ for symbols. We set m as a set of 1-hot vectors $\textbf{m}_1,...,\textbf{m}_T\ \in\ \left\{0,1\right\}^{|F|}$ for each source code token.
We represent the results of $E(Sym\left(Ident\left(CS\right)\right))$ as a set of vectors $\left\{\textbf{w}_{1}, ..., \textbf{w}_{T}\right\}$, which can be regarded as a learnable parameter for each token,
\begin{eqnarray}
\textbf{w}_{i}=\textbf{m}_i\textbf{F}
\end{eqnarray}
Since the context vector $\textbf{c}_t$ varies with time, the formation of context vector $c_t$ is as follows, 
\begin{equation}
\textbf{c}_t=\sum_{i=1}^{T}\alpha_{t,i}(\textbf{w}_{i}\ \textcircled{$\times$}\ {\textbf{e}_{3,i}})
\end{equation}
where $\textbf{e}_{3, i}$ is the hidden state located at the 3rd layer and $i$th position ($i=1,\dots,T$) in the encoder, $T$ is the input size. $\alpha_{t,i}$ is the weight term of $i$th location at step $t$ in the input sequence, which is used to tackle the situation when input piece is overlength. Then we can get a new form of $\textbf{y}_t$,
\begin{equation}
\textbf{y}_t=f_{d}(\textbf{c}_t, \textbf{d}_{3,t-1}, \textbf{d}_{2,t}, \textbf{y}_{t-1})
\end{equation}
$f_d(\cdot)$ is the decoder function. $\textbf{d}_{3,t}$ is the hidden state located at the 3rd layer and $t$th step ($t=1,\dots,K$) in the decoder. Here, we assume that the length of output is $K$. Instead of LSTM in~\cite{attention}, we take GRU~\cite{GRU} as basic unit in both $f_{e}\left(\cdot\right)$ and $f_{d}\left(\cdot\right)$.  Note that the weight term $\alpha_{t,i}$ is normalized to $\left[0,1\right]$ using a softmax function,
\begin{equation}
\alpha_{t,i}=\frac{\exp(\textbf{s}_{t,i})}{\sum_{i=1}^{T}\exp(\textbf{s}_{t,i})},
\end{equation}
where $\textbf{s}_{t,i}=score(\textbf{d}_{3,t-1}, \textbf{e}_{3,i})$ scores how well the inputs around position $i$ and the output at position $t$ match. As in~\cite{bahdanau2014neural}, we parametrize the score function $score(\cdot)$ as a feed-forward neural network which is jointly trained with all the other components of the proposed architecture.

\subsection{Ablation Study}
For a better demonstration of the effect of Code Attention, we make a naive ablation study about it. 
\begin{table}[t]
	\caption{Ablation study about effects of different parts in Code Attention. This table reports the BLEU-4 results of different combinations.}
	\label{ab}
	\centering
	\footnotesize 
	\begin{tabular}{c|c|c|c|c}
		& BLEU-4 & Ident & Token & Global Attention \\
		\hline
		\hline
		\textcolor{gray} {1} & 16.72 & w/o & w/o & w/o \\
		\textcolor{gray} {2} & 18.35 & w/ & w/o & w/o \\
		\textcolor{gray} {3} & 22.38 & w/ & w/ & \textcircled{+}\\
		\textcolor{gray} {4} & 24.62 & w/ & w/ &  \textcircled{$\times$}\\
		\hline
	\end{tabular}
\end{table}
For Table~\ref{ab}, we can get two interesting observations. First, comparing line 1 with line 2, we can see that even single {\em Identifier Ordering} have some effects. RNN with Identifier Ordering surpasses normal GRU-NN by 1.63, which reflects that stressing the order of different identifiers can be useful. The improvement can be reached when applying it on other methods based on neural machine translation. Second, \textcircled{$\times$} (line 4) surpasses \textcircled{$+$} (line 4) by 2.24, which precisely follows our intuition, that \textcircled{$\times$} amplifies the effects of tokens more efficiently than \textcircled{$+$} does.

\section{Experiments}
\label{ex}

We compared our Code Attention with several baseline methods on C2CGit dataset. The metrics contain both automatic and human evaluation.
\subsection{Baselines}

To evaluate the effectiveness of Code Attention, we compare different popular approaches from natural language and code translation, including CloCom, MOSES, LSTM-NN~\cite{srinivasan2016summarizing}, GRU-NN and Attention Model~\cite{attention}. All experiments are performed on C2CGit. It is worth noting that, for a better comparison, we improve the RNN structure in~\cite{vinyals2015grammar} to make it deeper and use GRU~\cite{GRU} units instead of LSTM proposed in the original paper, both of which help it become a strong baseline approach.
\begin{itemize}
	
	\item{\textbf{CloCom:}} This method raised by~\cite{wong2015clocom} leverages code clone detection to match code snippets with comment segments, which can't generate comment segments from any new code snippets. The code snippets must have similar ones in the database, then it can be annotated by existing comment segments. Hence, most code segments would fail to generate comments. CloCom also can be regarded as an information retrieval baseline.
	
	\item{\textbf{MOSES:}} This phase-based method~\cite{koehn2007moses} is popular in traditional statistical machine translation. It is usually used as a competitive baseline method in machine translation. We train a 4-gram language model using KenLM~\cite{heafield2011kenlm} to use MOSES.
	
	\item{\textbf{LSTM-NN:}} This method raised by~\cite{srinivasan2016summarizing} uses RNN networks to generate texts from source code. The parameters of LSTM-NN are set up according to~\cite{srinivasan2016summarizing}.
	
	\item{\textbf{GRU-NN:}} GRU-NN is a 3-layer RNN structure with GRU cells~\cite{cho2014learning}. Because this model has a contextual attention, it can be regarded as a strong baseline.
	
	\item {\textbf{Attention Model:}}~\cite{attention} proposed a new simple network architecture, the Transformer, based solely on attention mechanisms, dispensing with recurrence and convolutions
	entirely. The simple model achieves state-of-the-art results on various benchmarks in natural language processing.
\end{itemize}
\subsection{Automatic Evaluation}
We use BLEU~\cite{papineni2002bleu} and METEOR~\cite{banerjee2005meteor} as our automatic evaluation index. BLEU measures the average n-gram precision on a set of reference sentences. Most machine translation algorithms are evaluated by BLEU scores, which is a popular evaluation index.

METEOR is recall-oriented and measures how well the model captures content from the references in the output. \cite{denkowski2014meteor} argued that METEOR can be applied in any target language, and the translation of code snippets could be regarded as a kind of minority language. In Table~\ref{METEOR}, we report the factors impacting the METEOR score, e.g., precision, recall, f1, fMean and final score. 

In Table~\ref{BLEU}, BLEU scores for each of the methods for translating code snippets into comment segments in C2CGit, and since BLEU is calculated on n-grams, we report the BLEU scores when n takes different values.
\begin{table}[t]
	\footnotesize
	\caption{BLEU of Each Auto-Generated Comments Methods}
	\label{BLEU}
	\centering
	\setlength{\tabcolsep}{5pt} 
	\begin{tabular}{c|c|c|c|c}
		Methods & BLEU-1 & BLEU-2 & BLEU-3 & BLEU-4\\
		\hline
		\hline
		CloCom  &  25.31   &  18.67   &  16.06 &  14.13\\
		\hline 
		MOSES   &  45.20   &  21.70   &  13.78 &  9.54\\
		\hline 
		LSTM-NN  & 50.26   &  25.34   &  17.85 &  13.48\\
		\hline
		GRU-NN     &   58.69    &  30.93   &  21.42 &  16.72\\
		\hline
		Attention& 25.00 & 5.58 & 2.4 & 1.67 \\ 
		\hline
		Ours & \textbf{61.19}  &  \textbf{36.51}  &  \textbf{28.20} &  \textbf{24.62}\\
		\hline
	\end{tabular}
\end{table}
From Table~\ref{BLEU}, we can see that the BLEU scores of our approach are relatively high when compared with previous algorithms, which suggests Code Attention is suitable for translating source code into comment. 
Equipped with our Code Attention module, RNN gets the best results on BLEU-1 to BLEU-4 and surpass the original GRU-NN by a large margin, e.g., about 50\% on BLEU-4.

\begin{table}[t]
	\caption{METEOR of different comments generation models. \textbf{Precision}: the proportion of the matched n-grams	out of the total number of n-grams in the evaluated	translation; \textbf{Recall}: the proportion of the matched n-grams out of the total number of n-grams in the reference translation; \textbf{fMean}: a weighted combination of Precision and Recall; \textbf{Final Score}: fMean with penalty on short matches.}
	\label{METEOR}
	\centering
	\footnotesize
	\setlength{\tabcolsep}{5pt} 
	\begin{tabular}{c|c|c|c|c}
		Methods & Precision & Recall & fMean & Final Score \\
		\hline
		\hline
		CloCom  &   0.4068  & 0.2910  & 0.3571 & 0.1896 \\
		\hline 
		MOSES   & 0.3446 &   \textbf{0.3532}  &  0.3476  &   0.1618 \\
		\hline 
		LSTM-NN  &  0.4592   &   0.2090   &   0.3236  &  0.1532 \\
		\hline
		GRU-NN   &   0.5393   &   0.2397  &   0.3751 &    0.1785 \\
		\hline
		Attention   & 0.1369    &  0.0986   &  0.1205  &  0.0513 \\
		\hline
		Ours   & \textbf{0.5626}   &  0.2808  &  \textbf{0.4164}  &   \textbf{0.2051}  \\
		\hline
	\end{tabular}
\end{table}

Table~\ref{METEOR} shows the METEOR scores of each comments generation methods. The results are similar to those in Table~\ref{BLEU}. Our approach already outperforms other methods and it significantly improves the performance compared with GRU-NN in all evaluation indexes. Our approach surpasses GRU-NN by 0.027 (over 15\%) in Final Score. It suggests that our Code Attention module has an effect in both BLEU and METEOR scores. In METEOR score, MOSES gets the highest recall compared with other methods, because it always generates long sentences and the words in references would have a high probability to appear in the generated comments. In addition, in METEOR, the Final Score of CloCom is higher than MOSES and LSTM-NN, which is different from Table~\ref{BLEU} because CloCom can't generate comments for most code snippets, the length of comments generated by CloCom is very short. The final score of METEOR would consider penalty of length, so CloCom gets a higher score. 

Unexpectedly, Attention model achieves the worst performance among different models in both BLEU and METEOR, which implies that Attention Model might not have the ability to capture specific features of code snippets. We argue that the typical structure of RNN can be necessary to capture the long-term dependency in code which are not fully reflected in the position encoding method from Attention model~\cite{attention}.

\subsubsection{Human Evaluation}
Since automatic metrics do not always agree with actual quality of the results \cite{stent2005evaluating}, we perform human evaluation. This task refer to reading the Java code snippets and related comments, hence, we employed 5 workers with 5+ years Java experience to finish this task. The groudtruth would be read meanwhile rating the comments for eliminating prejudice. Each programmer rated the comments independently. The criterion would be shown in the following:
\begin{itemize}
	\item{\textbf{Understandability:}} we consider the fluency and grammar of generated comments. The programmers would score these comments according to the criterion shown by Table \ref{Criterion of fluency and grammar}. If programmers catch the meaning of code snippets in a short time, the scores of understandability would be high.
	\item{\textbf{Similarity:}} we should compare the generated comments with human written ones, which suggests what the models learn from the training set and the details are shown in Table \ref{Criterion between generated comments and human written}. This criterion measures the similarity between generated comments and human written.
	\item{\textbf{Interpretability:}} the connection between code and generated comments also should be considered. The detailed criterion is shown in Table \ref{Criterion between generated comments and code snippets}, which means the generated comments convey the meaning of code snippets. 
\end{itemize}

We randomly choose 220 pairs of code snippets and comment segments from the test set, and let programmers rate them according to above three evaluation. The automatic generated comments come from different methods would be shuffled before rating. The results would be shown in the following.

\begin{table}[!t]
	\caption{Criterion of Understandability}
	\label{Criterion of fluency and grammar}
	\centering
	\begin{tabular}{c|c}
		\hline
		Level & Meaning \\
		\hline
		5  &  Fluent, and grammatically correctly \\
		\hline
		4  &  Not fluent, and grammatically correctly \\
		\hline
		3  &  Grammatically incorrectly, but easy to understand \\ 
		\hline
		2  &  Grammatically incorrectly, and hard to understand \\
		\hline
		1  &  Meaningless \\
		\hline
	\end{tabular}
\end{table}

\begin{table}[t]
	\caption{Criterion of similarity between generated comments and human written}
	\label{Criterion between generated comments and human written}
	\centering
	\begin{tabular}{c|c}
		\hline
		Level & Meaning \\
		\hline
		5  &  \tabincell{c} {Generated comments are easier to understand \\ than  the human written} \\
		\hline
		4  & \tabincell{c}{ The meaning both generated and  human written \\ comments is same,  and the expression is same } \\
		\hline
		3  & \tabincell{c} { The meaning both generated and  \\ human written comments is same, \\ but the expression is different}  \\ 
		\hline
		2  &  \tabincell{c} {The meaning both generated and \\ human written comments is different, \\ but the generated comments  \\ express some information of code }\\
		\hline
		1  &  The generated comments is meaningless.\\
		\hline
	\end{tabular}
\end{table}

\begin{table}[t]
	\caption{Criterion of interpertability}
	\label{Criterion between generated comments and code snippets}
	\centering
	\begin{tabular}{c|c}
		\hline
		Level & Meaning \\
		\hline
		4  & \tabincell{c}{ The generated comments show the \\ high level meaning  in code snippets} \\
		\hline
		3  & \tabincell{c} {The generated comments only show \\ partial meaning  in code snippets.}  \\ 
		\hline
		2  &  \tabincell{c} {The generated comments only shows \\ some keywords in code snippets}\\
		\hline
		1  &  \tabincell{c}{There doesn't exist connection between \\ code snippets and generated comments.}\\
		\hline
	\end{tabular}
\end{table}

Table \ref{human} shows the human evaluation of all auto-generated comments methods from three aspects. The three aspects are understandability, similarity and interpretability. Our method gets the best performance in all aspects. It's suggested that our proposed method has an improvement than other methods in human evaluation. For details, we show the each human evaluation scores in the following.

\begin{figure}[!t]
	\centering
	{\includegraphics[width=1\columnwidth]{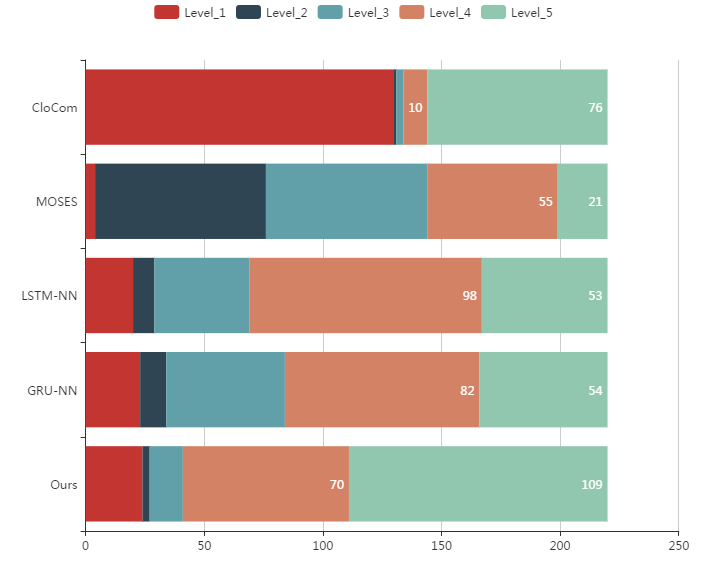} }
	\caption{Understandability distribution of each auto-generated comments methods}
	\label{Understandability}
\end{figure}

\begin{table}[t]
	\caption{The human evaluation of all methods}
	\label{human}
	\footnotesize
	\centering
	\begin{tabular}{c|c|c|c}
		\hline
		Methods & Understandability & Similarity & Interpretability  \\
		\hline
		CloCom  &   2.55  & 2.00 &  1.77  \\
		\hline 
		MOSES   & 3.08  &  2.84  & 2.60  \\
		\hline 
		LSTM-NN  &  3.70   &   2.96 &  2.39   \\
		\hline
		GRU-NN     &   3.60   & 3.27  & 2.76     \\
		\hline
		Ours & \textbf{4.08} & \textbf{3.36} &  \textbf{2.98}  \\ 
		\hline
	\end{tabular}
\end{table}
\textbf{Understandability.} From Figure~\ref{Understandability}, we are able to draw several conclusions. Firstly, our method, with maximum ratios of {\em good} comments (4 and 5 points), achieves the best results over other four approaches. Secondly, LSTM-NN and  GRU-NN obtain the most comments in the ``gray zones". The last phenomenon that draws much attention is ColCom has the worst performance in general, although it has 5 more points than GRU-NN and LSTM-NN . The reason might be the ColCom chooses the comments of similar code snippets as generated comments and these comments often have high quality. However, when facing many code snippets, ColCom can't generate enough appropriate comments.

\begin{figure}[t]
	\centering
	{\includegraphics[width=1\columnwidth]{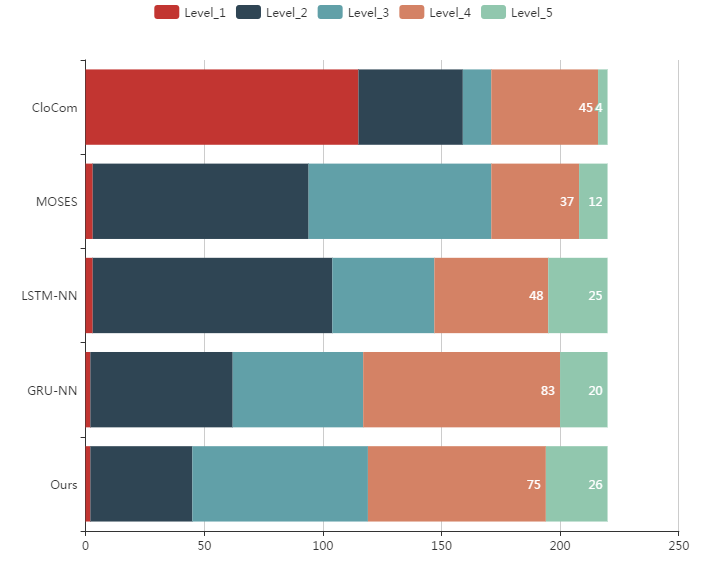} }
	\caption{Similarity distribution of each auto-generated comments methods}
	\label{Similarity}
\end{figure}

\textbf{Similarity.}
The results in Figure~\ref{Similarity} are nearly the same as those from Figure~\ref{Understandability}. We can easily tell that the ColCom has the least similar comments with ground-truth ones, which suggests that two code snippets might share many common words (because ColCom usually chooses the comments of similar code snippets) but the meaning of each could be different from the other. 

\textbf{Interpretability.}
When comes to the Interpretability, we can see that our method performs much better than other ones. The methods based on RNN architecture, e.g. LSTM-NN, GRU-NN, our method, much better than other methods. It's suggested that RNN architecture could catch the deep semantic not only literal meaning in code snippets.

\begin{figure}[t]
	\centering
	{\includegraphics[width=1\columnwidth]{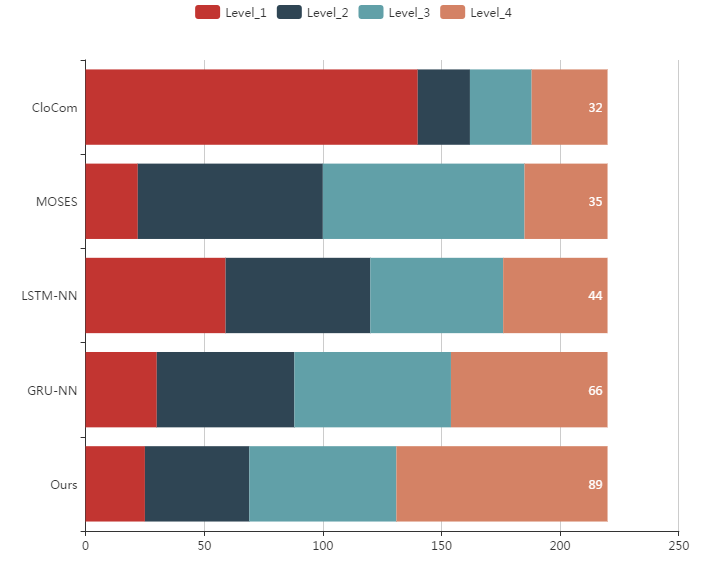} }
	\caption{Interpretability distribution of each auto-generated comments methods}
	\label{Interpretability}
\end{figure}

\subsubsection{Practical Comparison}
Table \ref{output_example} shows examples of the output generated by our models and other methods for code snippets in test set. Not all methods can generate meaningful sentences, suggesting the task is difficult and traditional methods having difficulties to achieve this goal. For the two examples, the comments translated by neutral networks are shorter than others and get the core meaning. Our method and GRU-NN regard the code snippets without condition or loop statements as the same. However, the generated comments are different with each other. It suggests that our proposed method can make the translation better though we only modify part of code snippets. MOSES generates longer comments than other methods, because it tends to make the length between source language and target language close, but the translation of source code does not match this assumption. LSTM-NN generates fluent sentences, which are shorter but information is less compared with our method. It's suggested that LSTM-NN can't catch the whole information  and it is not suitable for  code from real programming projects.

\begin{table*}[t]
	\caption{Two examples of code comments generated by different translation models.}
	\label{output_example}
	\centering
	\begin{tabular}{c|c}
		\hline code  &  \begin{lstlisting}[language=Java]
		private void createResolutionEditor(Composite control,
		IUpdatableControl updatable) {
		screenSizeGroup = new Group(control, SWT.NONE);
		screenSizeGroup.setText("Screen Size");
		screenSizeGroup.setLayoutData(new GridData(GridData.FILL_HORIZONTAL));\end{lstlisting} \\		
		\hline
		GroundTruth & the property key for horizontal screen size \\
		\hline
		ColCom & None \\
		\hline
		Moses &  \tabincell{c}{ create a new resolution control param control the control segment the segment size group \\ specified screen size group for the current screen size the size of the data is available} \\
		\hline
		LSTM-NN & creates a new instance of a size \\
		\hline
		GRU-NN&  the default button for the control \\
		\hline
		Attention & param the viewer to select the tree param the total number of elements to select \\
		\hline
		Ours &  create the control with the given size\\
		\hline
		\hline
		code & \begin{lstlisting}[language=Java]
		while (it.hasNext()) {  
		   EnsembleLibraryModel currentModel = (EnsembleLibraryModel) it.next();
		   m_ListModelsPanel.addModel(currentModel);
		   }\end{lstlisting} \\
		\hline
		GroundTruth & gets the model list file that holds the list of models in the ensemble library \\
		\hline
		ColCom & \tabincell{c}{ the library of models from which we can select our ensemble usually loaded from \\ a model list file mlf or model xml using the l command line option } \\
		\hline
		Moses  & adds a library model from the ensemble library that the list of models in the model  \\
		\hline
		LSTM-NN & get the current model \\
		\hline
		GRU-NN & this is the list of models from the list in the gui \\
		\hline
		Attention & the predicted value as a number regression object for every class attribute \\
		\hline
		Ours & gets the list file that holds the list of models in the ensemble library \\
		\hline
	\end{tabular}
\end{table*}

\subsection{Implementation Details}
For RNN architecture, as we have discussed above, we employed a 3-layer encoder-decoder architecture with a Code Attention module to model the joint conditional probability of the input and output sequences. 

\textbf{Adaptive learning rate}. The initial value of learning rate is 0.5. When step loss doesn't decrease after 3k iterations, the learning rate multiplies decay coefficient 0.99. Reducing the learning rate during the training helps avoid missing the lowest point. Meanwhile, large initial value can speed up the learning process. 

\textbf{Choose the right buckets}. We use buckets to deal with code snippets with various lengths. To get a good efficiency, we put every code snippet and its comment to a specific bucket, e.g., for a bucket sized $\left(40,15\right)$, the code snippet in it should be at most 40 words in length and its comment should be at most 15 words in length.
In our experiments, we found that bucket size has a great effect on the final result, and we employed a 10-fold cross-validation method to choose a good bucket size. After cross-validation, we choose the following buckets, $\left(40,15\right)$, $\left(55,20\right)$, $\left(70,40\right)$, $\left(220,60\right)$. 

We use stochastic gradient descent to optimize the network. In this network, the embedding size is 512 and the hidden unit size is 1024. Also, we have tried different sets of parameters. For example, 3-layer RNN is better than 2-layer and 4-layer RNNs, the 2-layer model has low scores while the 4-layer model’s score is only slightly higher than that of the 3-layer one, but its running time is much longer. Finally, it takes three days and about 90,000 iterations to finish the training stage of our model on one NVIDIA K80 GPU. We employ beam search in the inference.

\section{Conclusion}
\label{conclusion}

In this paper, we propose an attention module named Code Attention to utilize the specific features of code snippets, like identifiers and symbols. Code Attention contains 3 steps: Identifier Ordering, Token Encoding and Global Attention. Equipped with RNN, our model outperforms competitive baselines and gets the best performance on various metrics. Our results suggest generated comments would conform to the functional semantics of program, by explicitly modeling the structure of the code. In the future, we plan to implement AST tree into Code Attention and explore its effectiveness in more programming language.
\newpage

\bibliographystyle{unsrt}
\bibliography{ref}

\end{document}